\documentclass{egpubl}
\usepackage{egsgp2025}

 \SpecialIssuePaper         

\CGFccbyncnd

\usepackage[T1]{fontenc}
\usepackage{dfadobe}  

\usepackage{cite}  
\BibtexOrBiblatex
\electronicVersion
\PrintedOrElectronic
\ifpdf \usepackage[pdftex]{graphicx} \pdfcompresslevel=9
\else \usepackage[dvips]{graphicx} \fi

\usepackage{egweblnk} 

\usepackage{amssymb}
\usepackage{amsmath}
\usepackage{booktabs}
\usepackage{multirow}
\usepackage{adjustbox}
\usepackage{caption}
\usepackage{wrapfig}
\usepackage{xcolor}
\usepackage{subcaption}
\usepackage{wrapfig}
\usepackage{enumitem}

\newcommand{\change}[1]{{#1}}

\newcommand{\R}{\mathbb{R}}

\newcommand{\mM}{\mathcal{M}}
\newcommand{\mV}{\mathcal{V}}
\newcommand{\mF}{\mathcal{F}}
\newcommand{\ini}[1]{\widetilde{#1}}  
\newcommand{\refi}[1]{{#1}}        
\newcommand{\gt}[1]{{#1}^*}       

\newcommand{\tin}{\!\in\!}

\title[FRIDU: Functional Map Refinement with Guided Image Diffusion]%
      {FRIDU: Functional Map Refinement with Guided Image Diffusion}

\author[Avigail Cohen Rimon \& Or Litany  \& Mirela Ben-Chen]
{\parbox{\textwidth}{
        \centering 
        Avigail Cohen Rimon\orcid{0009-0000-7080-6091},
        Mirela Ben-Chen\orcid{0000-0002-1732-2327}
        and Or Litany\orcid{0000-0001-6700-7379} 
        }
        \\
{\parbox{\textwidth}{\centering Technion - Israel Institute of Technology\\
       }
}
}

\begin{document}

 \teaser{
  \includegraphics[width=0.95\linewidth]{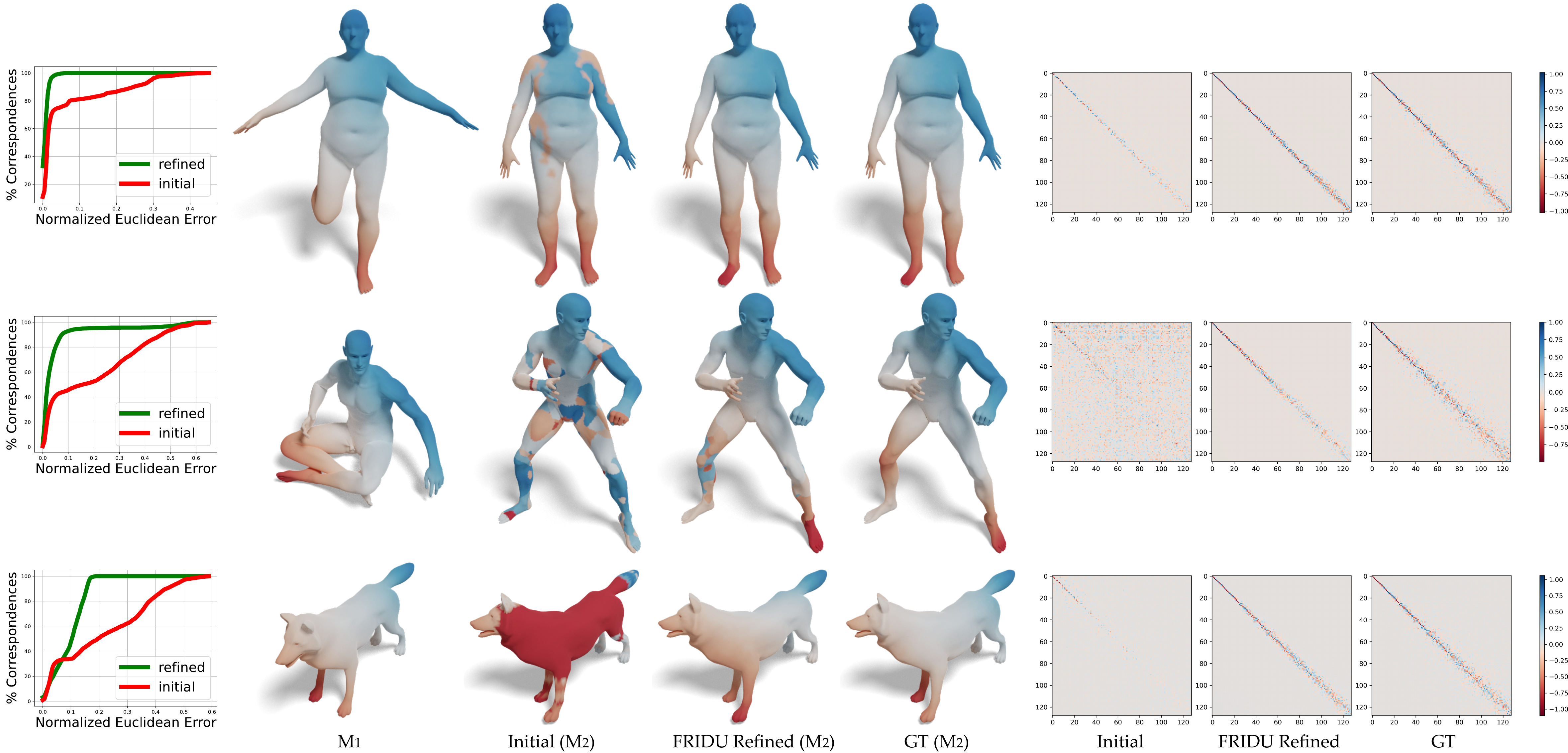}
  \centering
   \caption{FRIDU refines computed functional maps (Initial) via a guided image diffusion process.
   We show refinement results across different datasets and descriptor types: (top) FAUST with WKS, (middle) human TACO with SHOT, and (bottom) non-human TACO with WKS. Left: Error curves demonstrating consistent and significant improvement in correspondence quality. Middle: representative mappings from source to target shapes using the initial, FRIDU refined, and ground truth maps. Right: the corresponding functional map matrices.}
 \label{fig:teaser}
}

\maketitle
\begin{abstract}
We propose a novel approach for refining a given correspondence map between two shapes. A correspondence map represented as a \emph{functional map}, namely a change of basis matrix, can be additionally treated as a 2D image. With this perspective, we train an \emph{image diffusion model} directly in the space of functional maps, enabling it to generate accurate maps conditioned on an inaccurate initial map. The training is done purely in the functional space, and thus is highly efficient. At inference time, we use the pointwise map corresponding to the current functional map as \emph{guidance} during the diffusion process. The guidance can additionally encourage different functional map objectives, such as orthogonality and commutativity with the Laplace-Beltrami operator. We show that our approach is competitive with state-of-the-art methods of map refinement and that guided diffusion models provide a promising pathway to functional map processing. 

\printccsdesc   
\end{abstract}  
\section{Introduction}
Functional maps represent correspondences between shapes as a change of basis matrix~\cite{ovsjanikov2012functional,ovsjanikov2016computing}. This point of view has a myriad of benefits, such as efficiency and generality. Since their inception, multiple aspects of the functional maps computational pipeline have been extended and improved, for example by using different functional bases~\cite{wang2018steklov,hartwig2023elastic,bastian2024hybrid}, by incorporating them in a deep learning framework~\cite{litany2017deep,Halimi_2019_CVPR,sun2023spatiallyspectrallyconsistentdeep}, and by addressing issues such as partiality~\cite{rodola2017partial,litany2016non,litany2017fully,attaiki2021dpfm,bracha2024unsupervised} and smoothness~\cite{ren2018continuous,magnet2022smooth}, to mention just a few. 

However, not all functional maps represent a valid point to point (P2P) map. Hence, a major challenge is converting the functional map to a high quality P2P map for downstream applications. This task is known as functional map \emph{refinement},
and has been widely addressed in the literature, using both classical~\cite{ezuz2017deblurring,nogneng2018improved,ezuz2019reversible,melzi2019zoomout} and data-driven methods~\cite{magnet2024memory}.

Reconstructing a P2P map while refining the functional map is a strong regularizer which has been used in previous works, e.g. ZoomOut~\cite{melzi2019zoomout}. However, incorporating this step within a deep learning network is difficult, due to the large dimensions of the data involved, which depends on the dimensions of the input meshes, and the difficult non-linear constraints, essentially seeking a permutation. 

Image diffusion models~\cite{ho2020denoising,sohl2015deep,song2019generative,rombach2022high} have been hugely popular for image generation due to their high fidelity and flexibility. Initially primarily conditioned on text, these models have been adapted to support image-conditioned generation, allowing control via visual inputs for tasks such as style transfer, image-to-image translation, or super-resolution. In addition, guidance techniques~\cite{chung2022diffusion,song2023loss,bansal2023universal,yu2023freedom,he2023manifold} have been developed to steer the generative process at inference time using additional objectives -- enabling control over semantics, geometry, or alignment with external signals.

We propose that a \emph{conditioned image diffusion model with guidance} is a tool which is highly beneficial for functional map refinement. 
First, functional map matrices are treated as images, mapping the values to pixel values. Second, in the learning phase we work solely on the functional maps, learning to generate an "image" which represents a refined functional map conditioned on an "image" which represents a noisy initial functional map. The initial maps can be computed using different means, e.g., using descriptor correspondence. Finally, in the inference step, we use the input initial functional map as a condition and generate using the learned model a refined functional map.

This is highly efficient, as no P2P maps are required in the learning stage, since the ground truth functional maps are computed from the ground truth P2P maps in a preprocessing step. 
However, as is widely known~\cite{melzi2019zoomout}, restricting the output functional map to correspond to a P2P map is highly beneficial for the output map quality. Thus, in the inference step, we use a \emph{guidance objective} inspired by P2P reconstruction techniques to guide our output map, leading to refined functional maps that correspond to high quality P2P maps. 

We demonstrate that our approach is \change{flexible, accepting as input functional maps computed through various means } (e.g., different descriptors, deep feature extractors). Furthermore, we show that additional guidance objectives (e.g., orthogonality, or other functional map regularizations) can effectively improve the output map. \change{Notably, even when trained on} a very small dataset \change{our method successfully improves maps for shapes beyond the training distribution, as illustrated in Figure~\ref{fig:teaser} by an example involving a dataset distinctly different from the training set.} Finally, we show that our method compares favorably to the state of the art of functional map refinement approaches, both classical and data-driven.

\subsection{Related Work}
\paragraph*{Functional Map Refinement.}
Functional maps have been introduced by Ovsjanikov et al.~\cite{ovsjanikov2012functional}, and since then have been generalized in many ways~\cite{ovsjanikov2016computing,deng2022survey}. Here, we focus on map \emph{refinement}, namely improving an initial functional map, computed by some non-accurate means, e.g. from shape descriptors, such that an accurate point-to-point (P2P) map can be extracted from it. 

Already when introduced~\cite{ovsjanikov2012functional}, an iterative ICP algorithm in the spectral embedding space was proposed for extracting a P2P map. Later improvements included considering a smoothness assumption~\cite{ezuz2017deblurring}, namely that the pulled back Laplace-Beltrami eigenfunctions of one shape are \emph{in the span} of the Laplace-Beltrami eigenfunctions of the second shape. We also use this prior, though we leverage it as a \emph{guidance} term during the inference process. Thus we guide the functional map improvement using the P2P alignment prior. Among the many later refinement schemes, ZoomOut~\cite{melzi2019zoomout} and IMA~\cite{pai2021fast} stand out. ZoomOut uses the insight that alternatively upsampling the spectral dimension and projecting on the space of pointwise maps leads to better map recovery. IMA uses a connection to optimal transport to directly improve the functional map matrix. Recently, a differential version of ZoomOut was proposed~\cite{magnet2024memory}, which is incorporated within a network for learning shape correspondences. The refinement component there, however, has no learnable parameters.

As opposed to other map refinement methods, our approach has a few unique properties. First, to the best of our knowledge, it is the first method that \emph{learns} to refine functional maps from initial and ground truth maps, as opposed to computing refined maps from learned features. Second, the training phase is done fully on the functional map matrix (after a pre-computation of the initial and ground truth maps). Finally, we treat the functional maps as \emph{images} and leverage the powerful image diffusion models.


\paragraph*{Diffusion Models for Geometric Data.}
Diffusion models were originally introduced for high-fidelity 2D image generation~\cite{ho2020denoising,song2019generative,dhariwal2021diffusion,rombach2022high}, and have since been extended to a wide range of modalities. Their flexibility, stability, and controllable generation capabilities have recently made them attractive for 3D geometric data. This includes applications in 3D shape and scene synthesis, novel view synthesis, avatar modeling, and structure-aware feature extraction.

For geometric generation, diffusion models have been applied across a broad spectrum of representations. Early works explored point-based generation~\cite{luo2021diffusion,nichol2022point,melas2023pc2,wu2023sketch,zheng2024point}, while voxel and volumetric formulations enabled more structured geometry synthesis~\cite{hui2022neural,muller2023diffrf,tang2023volumediffusion}. Triplane-based models~\cite{chen2023single,shue20233d,wang2023rodin,zhang2024rodinhd,gupta20233dgen,long2024wonder3d} have proven effective for neural field generation and textured mesh synthesis. Direct mesh-based diffusion has also been explored~\cite{liu2023meshdiffusion}. More recently, 3D Gaussian Splatting (3DGS) has become a dominant representation. Several methods are trained directly on explicit 3D splat data~\cite{he2024gvgen,zhang2024gaussiancube,lan2023gaussian3diff,li2023gaussiandiffusion}, while a number of works have demonstrated that high-quality 3DGS can also be generated using only 2D supervision by leveraging differentiable rasterization and novel view consistency~\cite{peng2024lessonsplatsteacherguideddiffusion,szymanowicz2023viewset_diffusion,liu2024novelgsconsistentnovelviewdenoising,chen2024mvsplat360}. In parallel, 2D diffusion models have also been leveraged to create 3D assets via optimization-based lifting ~\cite{poole2022dreamfusion,tang2023dreamgaussian,lukoianov2024score}. Other approaches target novel view synthesis~\cite{watson2022novelviewsynthesisdiffusion,liu2023zero,sobol2024zero} and multimodal 3D outputs~\cite{xiang2024structured,wu2025amodal3r}, expanding the generative space to include conditional, occluded, or multi-view scenarios.

While diffusion models have been widely applied to 3D generation, their use for geometric understanding and shape correspondence remains limited. Some recent works have focused on extracting features from 2D renderings of meshes using diffusion backbones~\cite{dutt2024diffusion}, to decorate surfaces with semantic features. Others have repurposed 2D diffusion models fine-tuned for view synthesis to extract geometry-aware features~\cite{xu20243difftection,tang2023emergent}, which show strong view correspondence but are not used for shape matching or refinement. Je et al.~\cite{je2024robust} apply generative sampling via Riemannian Langevin Dynamics for robust symmetry detection, illustrating the broader use of probabilistic methods in geometric reasoning.

The most closely related and concurrent work is by Zhuravlev et al.~\cite{zhuravlev2025denoising}, who also use diffusion models to learn functional maps. Their approach predicts correspondences to a fixed template using a denoising pipeline trained on synthetic human shapes. In contrast, our method assumes a given (possibly noisy) functional map as input and \textit{refines} it between arbitrary shape pairs, with flexible plug-and-play guidance during inference.

\paragraph*{Guidance in Diffusion Models} 
Recent work has shown that diffusion models can be steered at inference time using guidance mechanisms that incorporate semantic or structural objectives. Classifier-based~\cite{dhariwal2021diffusion}, classifier-free~\cite{ho2022classifier}, and gradient-based plug-and-play methods~\cite{bansal2023universal,kawar2022denoising,wang2022zero,chung2022diffusion,lugmayr2022repaint,chung2022improving,graikos2022diffusion,he2023manifold,chung2023decomposed} allow external objectives to influence the generation process without retraining. These approaches have been used for semantic control, view consistency, and geometric constraints~\cite{peng2024lessonsplatsteacherguideddiffusion,szymanowicz2023viewset_diffusion,rempeluo2023tracepace}.

Our method follows this paradigm but applies guidance to functional map refinement, directing the diffusion process with geometric losses such as pointwise consistency, orthogonality, and Laplacian commutativity—objectives specific to spectral correspondence problems.

\subsection{Contributions}
Our main contribution is a reformulation of the functional map refinement problem as a conditional image generation problem. Specifically, we:
\begin{itemize}
\item Show that conditional image diffusion with guidance can refine functional maps generated by various sources. 
\item Provide a P2P guidance that efficiently improves the map at test time, as well as a variety of guidance mechanisms appropriate for different shape classes.
\item Demonstrate that our approach \change{compares favorably} to state of the art functional map refinement algorithms, both classical and deep.

\end{itemize}

\section{\change{FRIDU}}
\label{method-section}

\begin{figure*}
    \centering
    \includegraphics[width=1.05\linewidth]{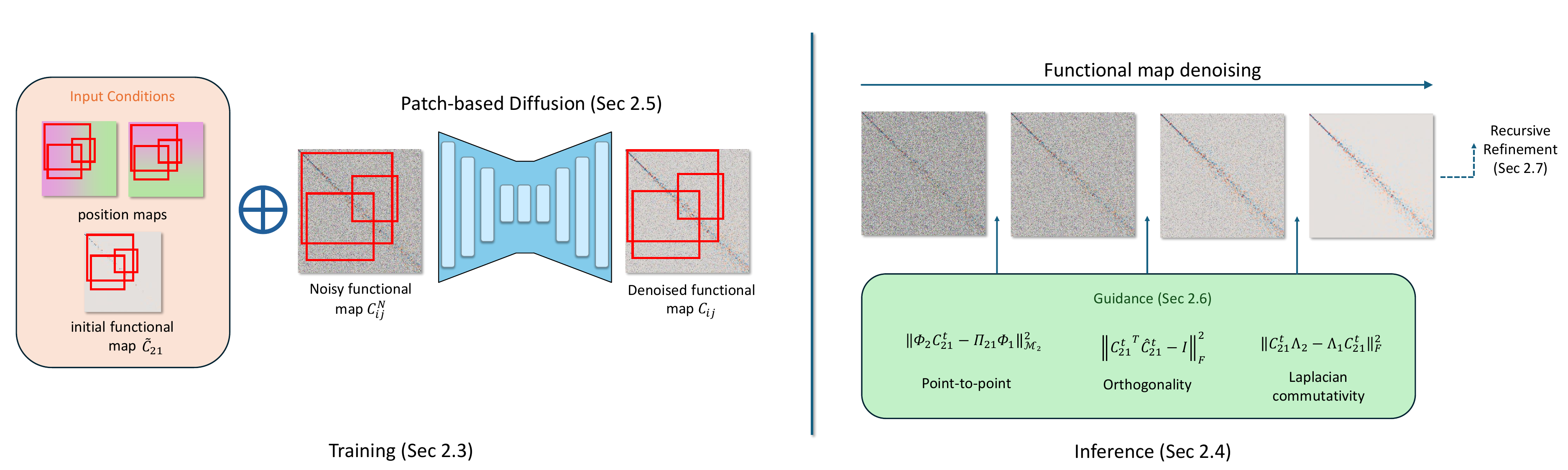}
    \caption{An illustration of our (left) training and (right) inference procedures. During training, our pipeline takes as input a random patch of a noisy functional map $\ini{C}_{ij}^N$, conditioned on a corresponding patch of an initial functional map $\ini{C}_{ij}$ and position maps, and outputs the denoised patch of the functional map $\boldsymbol{C}_{ij}$. At inference time, the patch covers the full-sized image, and we incorporate guidance at each denoising step, including point-to-point guidance and potentially additional regularizers, such as orthogonality and Laplacian commutativity.}
    \label{fig:pipeline_illustration}
\end{figure*}

\subsection{Notation}
We denote a pair of meshes by $\mM_j = (\mV_j,\mF_j)$, where $j=1..2$, and $\mV_j,\mF_j$ are the vertices and faces respectively, with $n_j = |\mV_j|, m_j = |\mF_j|$. The first $k_j$ eigenfunctions of the Laplace-Beltrami operator of the mesh $\mM_j$ are denoted by a matrix $\Phi_j \tin \R^{n_j \times k_j}$, and the corresponding diagonal matrix of $k_j$ eigenvalues is denoted by $\Lambda_j$. A point to point (P2P) map between the meshes is denoted by $T_{21}: \mM_2 \to \mM_1$, and maps vertices on $\mM_2$ to vertices on $\mM_1$. 

The P2P map can also be represented as a binary row stochastic matrix ${\Pi}_{21} \tin \{0,1\}^{n_2 \times n_1}$, where ${\Pi}_{21}(i,j) = 1$ if and only if $v_i \tin \mM_2$ is mapped to $v_j \tin \mM_1$. ${\Pi}_{21}$ maps \emph{indicator} functions defined on $\mV_1$ to \emph{indicator} (or zero) functions defined on $\mV_2$. 
A \emph{soft} P2P map $P_{21} \tin \R^{n_2 \times n_1}$ maps \emph{real} functions defined on $\mV_1$ to \emph{real} functions on $\mV_2$. A functional map $C_{21} \tin \R^{k_2 \times k_1}$ maps functions given in the spectral basis $\Phi_1$ to functions defined in the spectral basis  $\Phi_2$.

\subsection{Overview}
At inference time we are given an initial functional map $\ini{C}_{21}$ between two meshes $\mM_1, \mM_2$, as well as the corresponding functional bases $\Phi_1, \Phi_2$. We need to provide as output a refined functional map $\refi{C}_{21}$, and its corresponding refined P2P map $\refi{\Pi}_{21}$.

Towards that end we train a conditioned image diffusion model $d_\theta$ which generates our refined map $C_{21}$, conditioned on the initial map $\ini{C}_{21}$.
We train on maps $\ini{C}_{ij}$ between models $\mM_i, \mM_j$, from a dataset of models $\mathcal{D}_{\text{shapes}}$. These initial maps are given together with the ground truth P2P maps $\gt{\Pi}_{ij}$, from which we generate during pre-processing the corresponding ground truth functional maps $\gt{C}_{ij} = \Phi_i^\dagger \gt{\Pi}_{ij} \Phi_j$. The training is done solely in the functional space, namely only $\gt{C}_{ij}$ and $\ini{C}_{ij}$ are used during training.

The training and inference procedures are illustrated in Figure~\ref{fig:pipeline_illustration}.

\subsection{Training}
To simulate real-world scenarios where only noisy initial maps are available, we compute initial functional maps \( \ini{C}_{ij} \) using descriptor-based techniques (e.g., WKS~\cite{aubry2011wave} 
 or SHOT~\cite{tombari2010unique} descriptors). Thus, each training example consists of a pair \( (\ini{C}_{ij}, \gt{C}_{ij}) \), where \( \ini{C}_{ij} \) serves as the conditioning input and \( \gt{C}_{ij} \) is the supervision target.
We follow the denoising diffusion framework~\cite{ho2020denoising}, and adopt the EDM-DDPM++ formulation proposed by Karras et al.~\cite{karras2022elucidating}. In this formulation, the ground-truth functional map \( C^*_{ij} \) is corrupted using a diffusion process by adding Gaussian noise with a \textit{continuous} noise level \(\sigma\):
\begin{equation}
    {C}_{ij}^N = C^*_{ij} + \sigma \cdot \boldsymbol{\epsilon}, \quad \log \sigma \sim \mathcal{N}(\mu_p, \sigma_p^2) \quad \boldsymbol{\epsilon} \sim \mathcal{N}(\mathbf{0}, \mathbf{I})
\end{equation}
where \(\mu_p\) and \(\sigma_p\) are scalar parameters controlling the center and spread of the noise levels, and \( \boldsymbol{\epsilon} \in \mathbb{R}^{k_2 \times k_1} \) is a matrix of i.i.d.\ Gaussian noise with entries \( \epsilon_{a,b} \sim \mathcal{N}(0, 1) \), matching the dimensions of the functional map \( C^*_{ij} \).

Our model \( d_\theta \) is trained to directly recover the clean map from the noisy sample \( {C}_{ij}^N \), conditioned on the initial map and noise scale \(\sigma\).
We minimize the denoising objective:
\begin{equation}
    \mathbb{E}_{\substack{(\mathcal{M}_j, \mathcal{M}_i) \sim \mathcal{D}_{\text{shapes}}\\
    \boldsymbol{\epsilon} \sim \mathcal{N}(\mathbf{0}, \mathbf{I})\\
    \log \sigma \sim \mathcal{N}(\mu_p, \sigma_p^2)}} \left[
        \lambda(\sigma) \left\| \gt{C}_{ij} - d_\theta({C}_{ij}^N \mid \ini{C}_{ij}, \sigma) \right\|_2^2
    \right]
\end{equation}
where \(\lambda(\sigma)\) is a noise-dependent scalar weighting function. See~\cite{karras2022elucidating} for further details.

Training is performed entirely in the spectral (functional) domain, without requiring access to explicit P2P supervision. Once trained, the model can generate refined functional maps \( \refi{C}_{ij} \) from noise, guided by an initial estimate \( \ini{C}_{ij} \), which can then be decoded into dense correspondences using existing recovery techniques.

\subsection{Inference}
\begin{figure}[b]
    \centering
    \includegraphics[width=\linewidth]{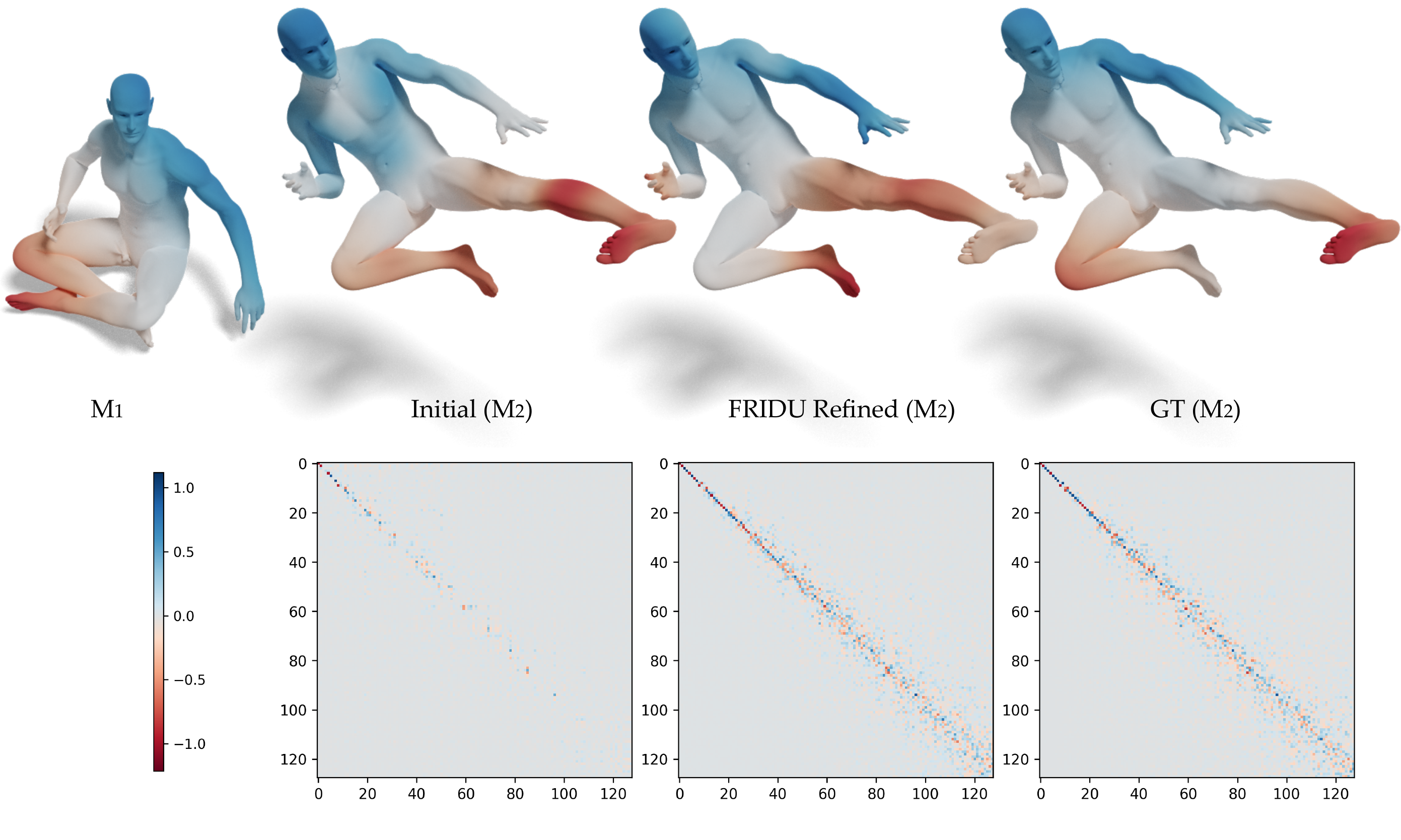}
    \caption{\textbf{Refinement without Guidance (WKS).} 
    We map a function $f_1 \in \mathbb{R}^{n_1}$ defined on $\mathcal{M}_1$ to $\mathcal{M}_2$, using the initial, FRIDU refined, and ground-truth \textbf{functional} maps. Here, we map the function $\Phi_1^\dagger f_1$ using the functional map matrices to get $\tilde{f}_2$, and show $\Phi_2 \tilde{f}_2$ on $\mM_2$. 
    Note that in this figure only, our refined map does \textit{not} include guidance in inference, in order to isolate and illustrate the refinement ability of the base model. The top row visualizes the source and mapped functions, and the bottom row shows the corresponding functional map matrices. We observe improvement in our refined map in both the mapped function and the functional map matrix.}
    \label{fig:michael-wks-fm}
\end{figure}
At inference time, we are given a new pair of shapes, their corresponding spectral bases \( \Phi_1, \Phi_2 \), and an initial functional map \( \ini{C}_{21} \). Our goal is to generate a refined functional map \( \refi{C}_{21} \), and recover accurate point-to-point correspondences $\refi{\Pi}_{21}$.

We initialize a noisy sample \( C_{21}^{t=T} \sim \mathcal{N}(0, \sigma_T^2 \mathbf{I}) \), and iteratively denoise it using our EDM-based model \( d_\theta \). At each step \( t = T, \dots, 1 \), the model predicts a cleaner version of the functional map corresponding to the current noise level \( \sigma_t \), and the sample is updated accordingly:

\begin{equation}
    C_{21}^{t-1} = C_{21}^t + (\sigma_{t-1}^2 - \sigma_t^2) \frac{d_\theta(C^t_{21} \mid \ini{C}_{21}, \sigma_t) - C^t_{21}}{\sigma_t^2}
\end{equation}

This update rule follows the deterministic sampling procedure proposed in~\cite{karras2022elucidating}. In practice, we use the full EDM sampler with noise perturbation and second-order correction for improved stability. The final output \( C^0_{21} \) is the refined functional map \( \refi{C}_{21} \).

\subsection{Efficient Training via Patch-based Diffusion}
Diffusion models typically require large amounts of training data, but in the case of functional maps, available datasets are relatively small. To improve training efficiency and data effectiveness, we adopt a \textit{patch-wise training strategy}~\cite{wang2023patch}.

Rather than denoising full functional maps, we train on smaller patches \( C^{(p)}_{ij} \in \mathbb{R}^{s \times s} \) randomly cropped from the ground-truth functional map \( C^*_{ij} \), with patch size \( s \) sampled stochastically or progressively. Each patch is denoised independently, conditioned on the corresponding region from the initial map \( \ini{C}_{ij} \), its spatial location \( (a, b) \), and the noise level. We concatenate two additional channels encoding the normalized position of each pixel within the full map to the patch input.

Patch locations \( (a, b) \) are sampled uniformly from the valid crop positions within the full functional map domain, denoted \( (a, b) \sim \mathcal{U}(\mathcal{R}) \). Patch sizes \( s \) are drawn from a predefined schedule \( p(s) \). 

The model is trained to minimize the patch-level denoising loss:
\begin{equation}
\mathbb{E}_{\substack{
(a, b) \sim \mathcal{U}(\mathcal{R})\\
s \sim p(s)\\
\log \sigma \sim \mathcal{N}(\mu_p, \sigma_p^2)\\
\boldsymbol{\epsilon} \sim \mathcal{N}(0, 1)^{s \times s}
}}
\left\| d_\theta\left(C^{(p)}_{ij} + \sigma \cdot \boldsymbol{\epsilon} \,\middle|\, \ini{C}^{(p)}_{ij}, a, b, s, \sigma \right) - C^{(p)}_{ij} \right\|_2^2
\end{equation}

To encourage the model to learn both local details and global structure, we vary the patch size during training and occasionally include full functional maps. Inference is performed on full maps without modification to the model architecture.

\subsection{Guidance} 
\begin{figure}
    \centering
    \includegraphics[width=\linewidth]{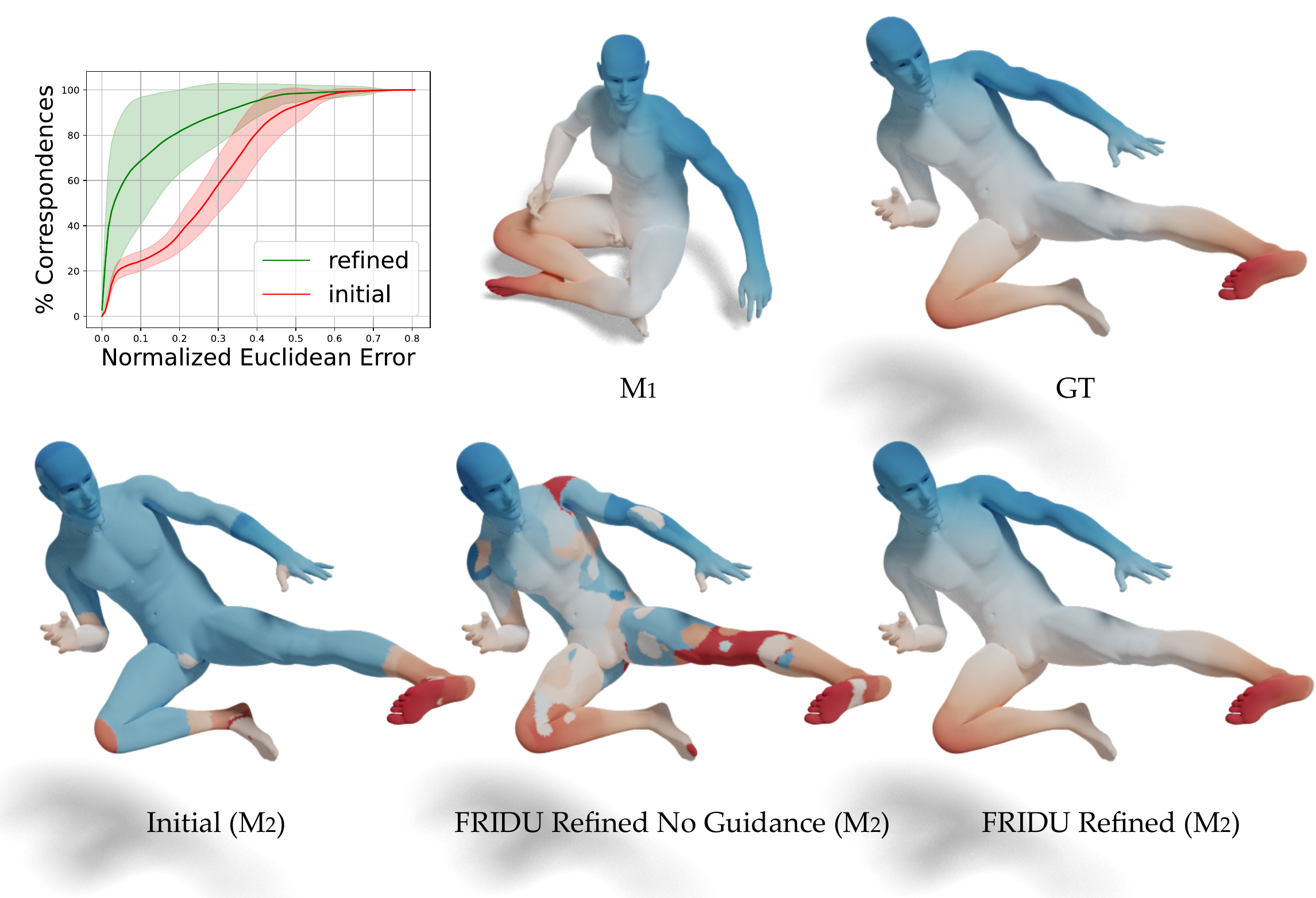}
    \caption{\textbf{Refinement with P2P Guidance (WKS).} 
    We show the \textbf{pointwise} mapping $\Pi_{21}$ extracted from the functional maps in Figure~\ref{fig:michael-wks-fm}, mapping a function $f_1 \in \mathbb{R}^{n_1}$ from $\mathcal{M}_1$ to $\mathcal{M}_2$ using $\Pi_{21} f_1$.
    We show the pointwise maps obtained \textit{without} guidance (center) and with P2P guidance (right). Note the improvement in the pointwise map extracted from the FRIDU refined functional map compared to the initial map, and the significant improvement when adding guidance. The plot shows the average normalized Euclidean error over the Michael dataset, for the initial and refined maps. The shaded region shows the standard deviation around the average.}
    \label{fig:michael-wks-pointwise}
\end{figure}
\paragraph*{Point to Point Map Extraction.}
The conditional diffusion model produces plausible functional maps $\refi{C}_{21}$ that resemble the training distribution (see Figure~\ref{fig:michael-wks-fm}). 
However, we also need to compute the P2P map $\refi{\Pi}_{21}$. \change{In general, since the functional maps are given in a reduced spectral basis (and thus are \emph{smoothed} versions of the P2P maps), there are many possible corresponding P2P maps (see e.g. the discussion in Ezuz et al.~\shortcite{ezuz2017deblurring}). Hence, some regularization is required to solve this ambiguity.} A minimal additional regularizing requirement, is that smooth functions on $\mM_1$ are smooth on $\mM_2$ after the map, which is formalized using \( \Pi_{21} \Phi_1 \in \text{span}(\Phi_2) \). This combined with the requirement that $\|C_{21} - \Phi_2^\dagger \Pi_{21} \Phi_1 \| $, leads to the optimization problem~\cite{ezuz2017deblurring}:
\begin{equation}
\label{eq:ezuz17}
 \Pi_{21}(C_{21}) = \operatorname*{arg\,min}_{\Pi_{21}\in \mathcal{P}_{21}} \left\| \Phi_2 C_{21} - \Pi_{21} \Phi_1 \right\|^2_{\mM_2},
\end{equation}

where \( \mathcal{P}_{21} \) is the set of valid maps, i.e. binary row stochastic matrices of dimension $n_2 \times n_1 $. This is easily optimized using a nearest neighbor search between the rows of $\Phi_2 C_{21}$ and the rows of $\Phi_1$~\cite[Sec 4.2]{ezuz2017deblurring}.

\paragraph*{Point to Point Guidance.}
Extracting the P2P map that corresponds to a refined functional map does not, however, result in a high quality map (see Figure~\ref{fig:michael-wks-pointwise}). Thus, at inference, we would like to incorporate a structural constraint that the computed functional map corresponds to a P2P map.  

Diffusion models allow such structure to be injected via \emph{guidance} at inference time~\cite{chung2022diffusion,song2023loss,bansal2023universal}, without requiring retraining. 
In our case, we wish to guide the model toward refined maps that produce accurate point-to-point correspondences, hence a natural guidance loss would be to use Eq.~\eqref{eq:ezuz17}:
\begin{equation}
    \|\Phi_2 C_{21}^t - \Pi_{21}(C_{21}^{t})\Phi_1\|^2_{\mM_2}.
\end{equation}

However, directly differentiating through the solution of this optimization is computationally expensive due to the nearest neighbor computation for $n_2$ points in $\R^{k_1}$ which is required for computing $\Pi_{21}$.

Instead, at each step, we solve for \( \Pi_{21} \)  but stop gradients through it — treating it as a fixed prediction. Hence, our P2P loss for guidance is:
\begin{equation}
\mathcal{L}_{\text{P2Pg}}(C_{21}^t) =    \|\Phi_2C_{21}^t - \Pi_{21}\Phi_1\|^2_{\mM_2}.
\end{equation}

Although \( \Pi_{21} \) is treated as constant during backpropagation, the iterative nature of the denoising process ensures that updates to \( C_{21} \) affect future \( \Pi_{21} \) values. This indirect influence allows the guidance signal to propagate across timesteps, effectively steering the generation toward spectrally and geometrically consistent solutions. Our final P2P map is $\Pi_{21}(C_{21}^0)$.
Figure~\ref{fig:michael-wks-pointwise} shows the resulting P2P map when using this guidance loss. Note the considerable improvement compared to inference without P2P guidance.

\paragraph*{Task-dependent Guidance.}
This approach to guidance is general and can accommodate a wide variety of loss functions, including enabling adaptation to new tasks or assumptions at inference time.
Such flexibility is a key advantage of diffusion-based refinement, enabling controlled generation without compromising generality.
In the experimental section (Table~\ref{table:comparisons}), we report results where two classical functional map regularizers are added to the guidance: the orthogonality constraint 
$L_{\text{orth}}(\boldsymbol{C_{21}}) = \| \boldsymbol{C_{21}}^T \boldsymbol{C_{21}} - \boldsymbol{I} \|_F^2$
\change{(promoting area preserving maps)} and the Laplacian commutativity term $L_{\Delta}(C_{21}) = \|C_{21}\Lambda_1 - \Lambda_2 C_{21}\|_F^2$ \change{(promoting isometric maps)}, see e.g.~\cite{cao2022unsupervised, ren2019structured}.

\subsection{Recursive Refinement}
\label{subssec:rec-ref}
Following the logic of refining a given noisy functional map, we conduct an experiment in which the model is recursively fed its own denoised output during \textit{inference} to assess whether performance continues to improve. In our experiments, we observe that a single recursive iteration typically enhances accuracy, but applying more than one iteration leads to degradation. We report results for adding one recursive iteration during inference in Table~\ref{table:comparisons}. Note that this additional refinement entails a computational trade-off.

\section{Experimental Results}
The following section is organized as follows:
In Section~\ref{subsec:analysis}, we evaluate our model's performance when conditioned on initial functional maps computed different descriptors, and demonstrate its dataset generalization capabilities.
In Section~\ref{subsec:comparisons}, we compare our method to existing shape matching approaches, and in Section~\ref{subsec:ablation}, we present an ablation study exploring zero-shot condition generalization and guidance parameters.

\subsection{Functional Map Refinement}
\label{subsec:analysis}
\label{subsec:shot-based-fm}
\begin{figure}
    \centering
    \includegraphics[width=\linewidth]{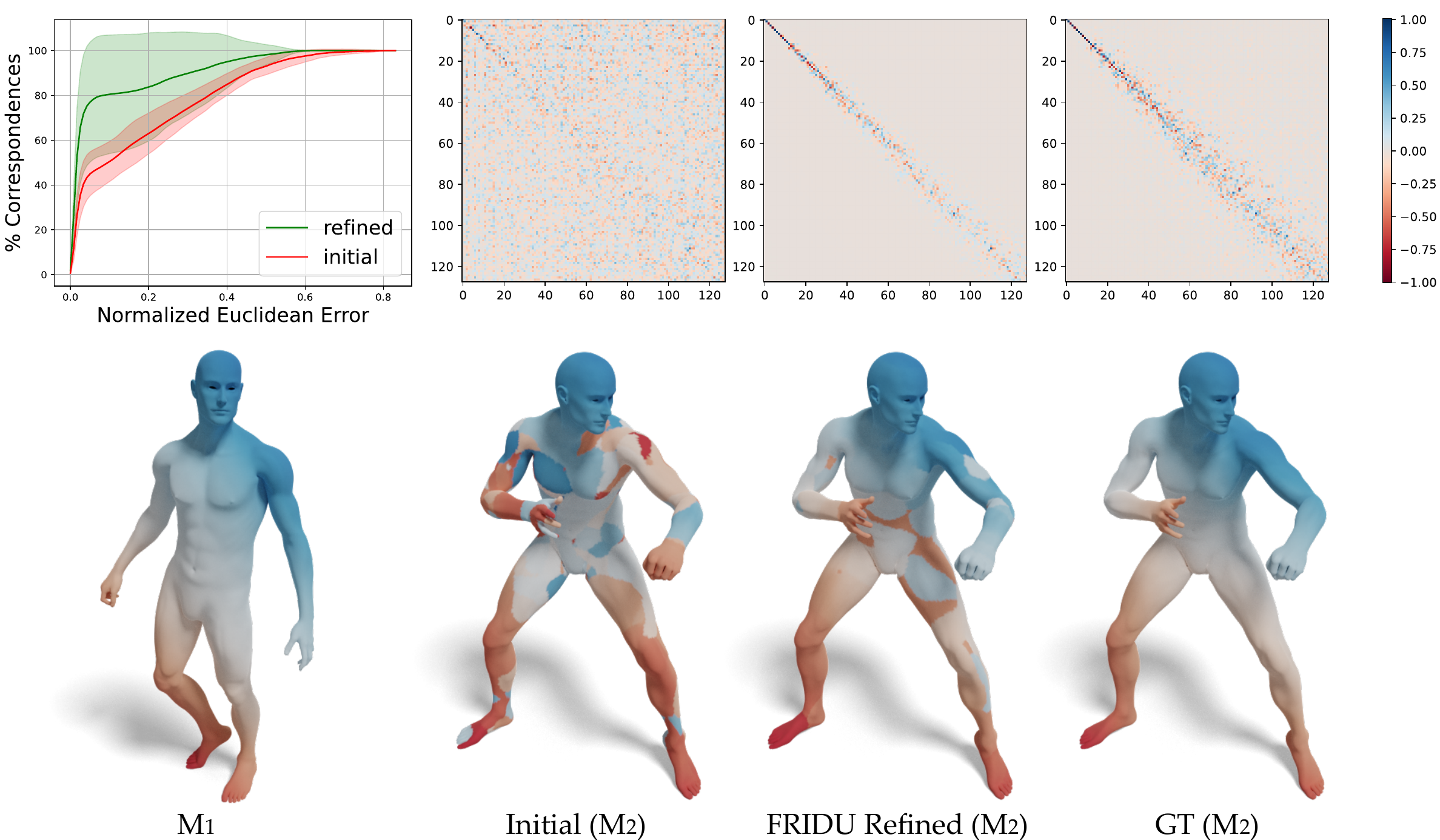}
    \caption{\textbf{Functional Map Refinement (SHOT).} We show the performance of the pointwise mapping extracted from our refined map alongside the initial and ground-truth mappings. We additionally show the corresponding functional map matrices.
    Note the noisy appearance of the SHOT-based initial map.
    The plot shows the normalized Euclidean error over the Michael dataset, where our refined maps consistently outperform the initial maps.}
    \label{fig:michael-shot}
\end{figure}
To demonstrate the effectiveness of our refinement model, we apply it to initial functional maps computed from both WKS ~\cite{aubry2011wave} and SHOT \cite{salti2014shot} descriptors in a classical way; see details in Appendix~\ref{sec:init_fmap}.

\paragraph*{Data.} We use the \textit{Michael} meshes from the TACO~\cite{pedico2024taco} dataset, which provides remeshed versions of different figures in various poses, along with ground-truth correspondences within each figure. Each mesh contains approximately 50\change{K} vertices. The dataset includes 190 ground-truth correspondences between the 20 \textit{Michael} shapes. We split the dataset into training and testing sets using a 90:10 ratio.

\paragraph*{Initial map from WKS.}
Figure~\ref{fig:michael-wks-fm} shows an example of mapping a function $f_1 \in \mathbb{R}^{n_1}$ defined on $\mathcal{M}_1$ to function on $\mathcal{M}_2$.  
The upper row shows the function on $\mathcal{M}_1$ and its mapping to $\mathcal{M}_2$ using (i) the initial map $\ini{C}_{21}$, (ii) our refined map $\boldsymbol{C}_{21}$, and (iii) the ground-truth map $\gt{C}_{21}$. Here we map $\Phi_1^\dagger f_1$ using the functional map matrix to get $\tilde{f}_2$, and show $\Phi_2 \tilde{f}_2$ on $\mM_2$. 
Note that only in this figure our refined version doesn't include guidance during inference, to demonstrate the ability of the base-model refinement abilities beside guidance.
The bottom row presents the corresponding functional map matrices. We observe that our refined map is closer to the ground truth than the initial one, successfully resolving artifacts. Some symmetry flips remain, which we discuss later in this subsection.

Figure~\ref{fig:michael-wks-pointwise} illustrates the pointwise map extracted from the corresponding functional maps, comparing our refined mapping obtained with and without incorporating P2P guidance during inference.
The bottom row shows the pointwise mapping of \( f_1 \) to \( \mathcal{M}_2 \) using (i) the initial extracted map \( \ini{\Pi}_{21}  \), (ii) our refined extracted map \textit{without} guidance, and (iii) our refined extracted map \textit{with} guidance \( \boldsymbol{\Pi}_{21} \).
The top row shows \( f_1 \) on \( \mathcal{M}_1 \) and its ground-truth mapping on \( \mathcal{M}_2 \), obtained using \( \gt{\Pi}_{21} \). Here all the pointwise maps are in $\mathbb{R}^{n_2 \times n_1}$, and thus the mapping is done using, e.g., $\Pi_{21} f_1$.
We additionally show the normalized Euclidean error over the test set for the initial and refined (\textit{with} guidance) maps. We observe that our approach improves performance both qualitatively and quantitatively.

From now on, whenever we refer to our refined mapping, we mean the version refined \emph{with P2P guidance} during inference.

\paragraph*{Initial Map from SHOT.}
Figure~\ref{fig:michael-shot} presents the performance of our model when conditioned on SHOT-based functional maps. 
The bottom row shows \( f_1 \) on \( \mathcal{M}_1 \) and the (i) initial, (ii) our refined, and (iii) ground-truth pointwise mappings. 
The top row shows the corresponding functional map matrices and the normalized Euclidean error over the test dataset for both the initial and our refined maps.
Note that both the initial pointwise mapping and functional map appear quite noisy, while our refined version resolves many of the artifacts, despite still having some. 
We observe that although our refined maps exhibit higher error and variance compared to the WKS-based maps experiment, our refined maps still outperforms the initial ones. 
See an additional example of SHOT-based map denoising in Figure~\ref{fig:teaser}.

\begin{figure}
    \centering
    \includegraphics[width=\linewidth]{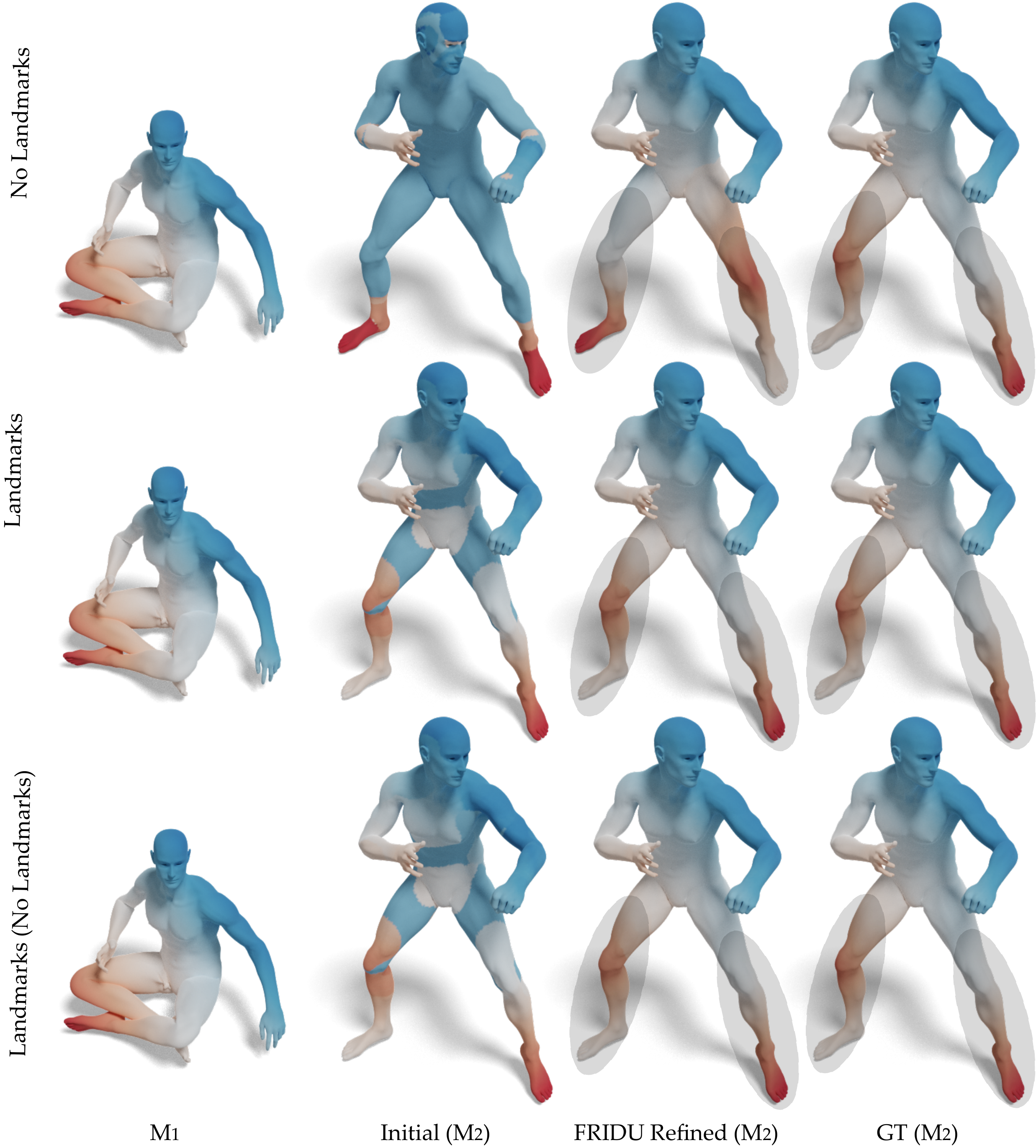}
    \caption{\textbf{Landmarks.} We show results for three settings: (top) training and evaluating on WKS-based initial maps, (center) training and evaluating on WKS-based initial maps with landmarks, and (bottom) training on WKS-based initial maps and evaluating on WKS-based initial maps with landmarks. 
    We note that symmetry flips are resolved in cases where the model is conditioned on an initial map with landmarks, even without training on such maps (see the shaded regions).}
    \label{fig:michael-wks-landmarks}
\end{figure}
\begin{figure}[t]
    \centering
    \includegraphics[width=0.6\linewidth]{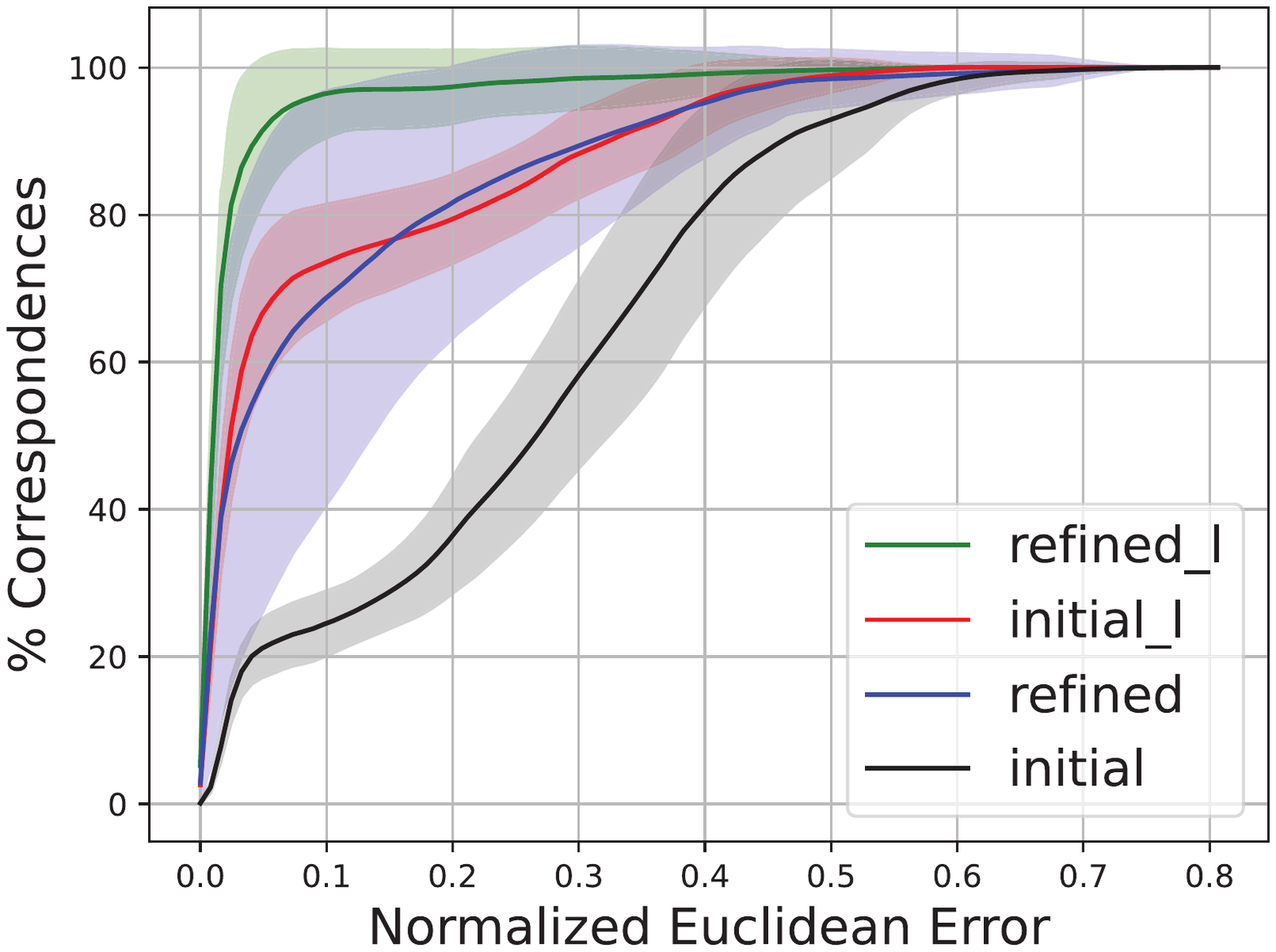}
    \caption{\textbf{Landmarks.} The normalized Euclidean error over the Michael dataset for the regular ("refined") and landmarks-constrained ("refined\_l") settings. For each setting, we also report the error of the initial maps ("initial" and "initial\_l", respectively). 
    Note that incorporating landmark constraints improves the accuracy of both the initial and refined maps, with our refined maps consistently outperforming the initial maps in both settings.}
    \label{fig:michael-wks-landmarks-err}
\end{figure}

\paragraph*{Landmark Constraints.}
If a sparse set of corresponding landmarks is available, it can be easily incorporated in our framework, by adding functional constraints based on these landmarks (e.g. as implemented in PyFM~\cite{magnet2021pyfm}) when computing the functional map. Landmarks constraints are useful for improving the quality of the correspondence, especially in the presence of intrinsic symmetries, which lead to non-unique eigendecomposition of the Laplace-Beltrami operator. When only symmetry-invariant descriptors (such as the WKS) are used, symmetry cannot be disambiguated without an additional step of aligning the LB eigenfunctions (see e.g. ~\cite{zhuravlev2025denoising}), or adding information that disambiguates the symmetry, such as landmarks.

To demonstrate this, we mark $5$ landmarks (head, two hands, and two legs) on both source and target shapes and incorporate these landmark constraints into the initial functional map computation. 
Figure~\ref{fig:michael-wks-landmarks} compares three settings: training and evaluating using the WKS-only initial maps (top row), training and evaluating using initial maps computed with landmarks (middle row), and training using the WKS-only initial maps and evaluating using the initial maps with landmarks (bottom row).
From left to right, for each setting we show \( f_1 \) on \( \mathcal{M}_1 \), and its pointwise mapping extracted from the (i) initial, (ii) our refined, and (iii) ground-truth mappings. 
We observe that the symmetric flip of the legs (see the shaded oval regions) present in the top row is resolved whenever conditioning on initial maps with landmarks (middle and bottom rows).

Figure~\ref{fig:michael-wks-landmarks-err} presents the normalized Euclidean error over the test dataset for both the regular and landmark-constrained settings.
For each setting, we show the error for the initial and refined extracted pointwise maps. We observe that the landmark-constrained setting improves both the initial and refined map errors compared to the regular setting, with our refined maps outperforming the initial maps in both cases.

\paragraph*{Dataset Generalization.}
To evaluate dataset generalization, we apply the model trained on the \textit{Michael} dataset, conditioned on a WKS-based initial functional map, to shape pairs of human figures outside this dataset, as well as to a non-human figure from the TACO dataset. 
The top row of Figure~\ref{fig:teaser} shows results for mapping a function between a pair from the FAUST dataset, while the bottom row shows mapping results for a pair of \textit{wolf} shapes from TACO. Note we use the remeshed version of FAUST.  
From left to right, each row shows (i) the normalized Euclidean error graph of the initial and FRIDU refined maps, (ii) the source and pointwise mapping of \( f_1 \) using the initial, FRIDU refined, and ground-truth mappings, and (iii) the corresponding functional map matrices. \change{The average initial and final errors for the top row (human) are $0.19$ and $0.05$, respectively, and for the third row (wolf) are $0.21$ and $0.03$, respectively.}
Note that despite being trained on the small \emph{Michael} dataset, our model generalizes well and significantly improves the quality of the initial maps—both quantitatively and qualitatively—on these unseen shape categories.

\subsection{Shape Matching Comparisons}
\label{subsec:comparisons}
\small
\begin{table}[b]
\centering
\begin{adjustbox}{max width=\linewidth}
\begin{tabular}{l|ccc|ccc}
\toprule
\multirow{2}{*}{} & \multicolumn{3}{c|}{\textbf{Train: F}} & \multicolumn{3}{c}{\textbf{Train: F+S}} \\
\cmidrule(r){2-4} \cmidrule(r){5-7}
\textbf{Method} & F & S & S19 & F & S & S19 \\
\midrule
BCICP \cite{ren2018continuous} & 6.4 & - & - & - & - & - \\
ZoomOut \cite{melzi2019zoomout} & 6.1 & - & - & - & - & - \\
SmoothShells \cite{eisenberger2020smooth} & \textbf{2.5} & - & - & - & - & - \\
DiscreteOp \cite{ren2021discrete} & 5.6 & - & - & - & - & - \\
\midrule
ACSCNN \cite{li2020shape} & 2.7 & \textbf{8.4} & - & - & - & - \\
TransMatch \cite{trappolini2021shape} & \textbf{1.7} & 30.4 & 14.5 & \textbf{1.6} & 11.7 & 10.9 \\
\midrule
GeomFmaps \cite{donati2020deep} & 3.5 & 4.8 & 8.5 & 3.5 & 4.4 & 7.1 \\
Deep Shells \cite{eisenberger2020deep} & 1.7 & 5.4 & 27.4 & 1.6 & 2.4 & 21.1 \\
NeuroMorph \cite{eisenberger2021neuromorph} & 8.5 & 28.5 & 26.3 & 9.1 & 27.3 & 25.3 \\
DUO-FMNet \cite{donati2022deep} & 2.5 & 4.2 & 6.4 & 2.5 & 4.3 & 6.4 \\
UDMSM \cite{cao2022unsupervised} & \textbf{1.5} & 7.3 & 21.5 & 1.7 & 3.2 & 17.8 \\
ULRSSM \cite{cao2023unsupervised} & 1.6 & 6.4 & 14.5 & \textbf{1.5} & \textbf{2.0} & 7.9 \\
ULRSSM + fine-tune \cite{cao2023unsupervised} & 1.6 & \textbf{2.2} & 5.7 & 1.6 & 2.1 & 4.6 \\
AttentiveFMaps \cite{li2022learning} & 1.9 & 2.6 & 5.8 & 1.9 & 2.3 & 6.3 \\
ConsistentFMaps \cite{sun2023spatially} & 2.3 & 2.6 & \textbf{3.8} & 2.2 & 2.3 & 4.3 \\
\midrule
DiffZO \cite{magnet2024memory} & 1.9 & 2.9 & 4.2 & 1.9 & 2.3 & \textbf{3.6} \\
\midrule
Initial & 2.5 & 6.9 & 12.4 & 2.1 & 2.7 & 8.6 \\
Ours & 1.7 & 4.1 & 9.4 & 1.5 & 2.1 & 7.1 \\
 + Recursive Refinement (Sec.~\ref{subssec:rec-ref}) & 1.6 & 4.1 & 9.3 & \textbf{1.2} & \textbf{1.9} & 6.6 \\
 + Orthogonality Guidance & \textbf{1.5} & \textbf{2.7} & \textbf{7.3} & 1.4 & 2.0 & \textbf{6.5} \\
 + Laplacian Commutativity Guidance & 1.7 & 4.4 & 9.8 & 1.5 & 2.0 & 7.3 \\
\bottomrule
\end{tabular}
\end{adjustbox}
\caption{Mean geodesic errors (×100) when training and testing on the FAUST, SCAPE and SHREC19 datasets. Best result within each method category (axiomatic, supervised, and unsupervised methods) is shown in bold. We consider DiffZO as part of the unsupervised category.
The axiomatic and supervised methods are from \cite{sun2023spatially} and the unsupervised methods are from \cite{magnet2024memory}.
}
\label{table:comparisons}
\end{table}
\begin{figure}[b]
    \centering
    \includegraphics[width=\linewidth]{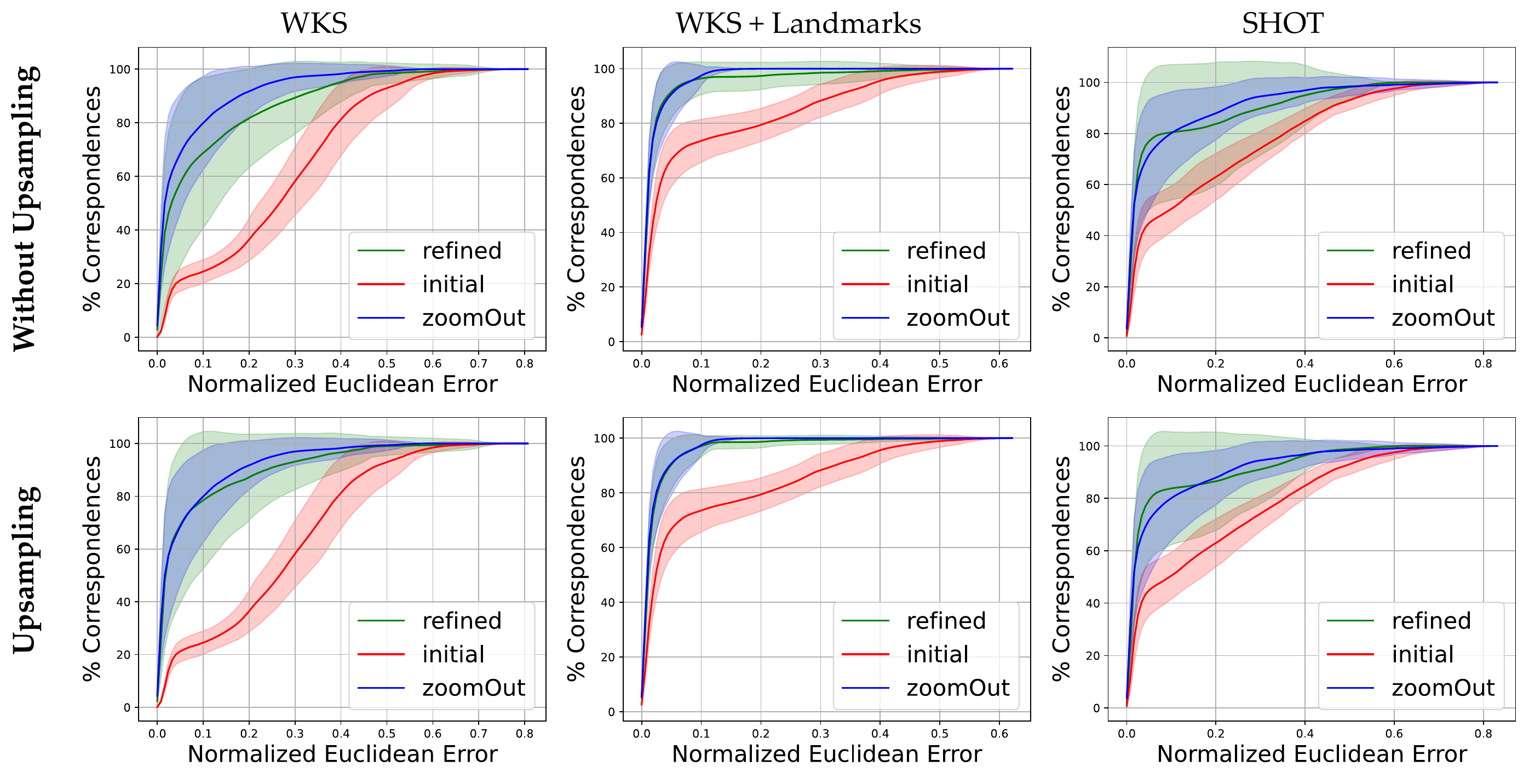}
    \caption{\textbf{Comparison to ZoomOut.} For the three initial map computation methods presented in Section~\ref{subsec:analysis}, we show a comparison to the ZoomOut axiomatic refinement algorithm \cite{melzi2019zoomout} on the Michael dataset. We show results with and without incorporating spectral upsampling during inference of our method. We observe that when incorporating upsampling, our method is comparable to ZoomOut.}
    \label{fig:zoomout-comparison}
\end{figure}

Since deep learning-based functional map \textit{refinement} is less established, we position our method within the broader context of deep learning for shape \textit{matching}. Most existing approaches consist of a learned feature extractor followed by functional map computation, typically optimized with geometric regularizers. Our method is compatible with both supervised and unsupervised pipelines, as it operates independently of how the initial functional map is obtained. To isolate the effect of the refinement step, we use as input the functional map produced by the feature extractor of a state-of-the-art method, DiffZO. Notably, DiffZO includes a built-in refinement stage, allowing for a direct comparison between their refinement and ours under the same initial conditions. This setup also demonstrates our model’s ability to refine functional maps originating from different sources.

We train two models — one on FAUST \cite{bogo2014faust} and the other on FAUST + SCAPE \cite{anguelov2005scape} —and evaluate them on FAUST, SCAPE and SHREC’19 \cite{melzi2019shrec} datasets, using the remeshed version \cite{ren2018continuous} of each dataset. \change{We used the same train/test split as in the DiffZo experiments. Specifically, for the FAUST remeshed dataset we used $80$ pairs for training and $20$ for testing, and for the SCAPE remehed dataset we used $51$ for training and $20$ for testing. The SHREC19 dataset was only used for testing. }

Table~\ref{table:comparisons} reports the geodesic error ×100 for axiomatic, supervised, and unsupervised methods.
Note that the table presents results for our base pipeline, as well as for three variants: one using a recursive refinement operator (as described in \change{Sec.} \ref{subssec:rec-ref}), and another incorporating \change{two} common functional map regularizers into the guidance. We also report the error of the initial functional map given as input to our model.
Our method consistently improves the initial mapping and outperforms DiffZO in \change{intra}-dataset evaluations, while maintaining comparable results in cross-dataset evaluations relative to supervised methods. We emphasize that our primary point of comparison is DiffZO, as our model is conditioned on maps generated using its feature extractor.

Among the axiomatic refinement methods listed, SmoothShells \cite{eisenberger2020smooth} requires access to the full shape geometry during the refinement process — a requirement our method avoids.
The DiscreteOp approach \cite{ren2021discrete} is tailored to minimizing a given functional map energy while keeping the maps \textit{proper}. In contrast, our method does not rely on functional map objectives and except the spectral guidance operates work purely in image space. However, we show that regularization can be incorporated into the guidance when needed, without additional training.
The method most closely related to ours is ZoomOut \cite{melzi2019zoomout}, which has been shown to outperform BCICP \cite{ren2018continuous} in both accuracy and runtime.

\begin{figure*}
    \centering
    \includegraphics[width=\linewidth]{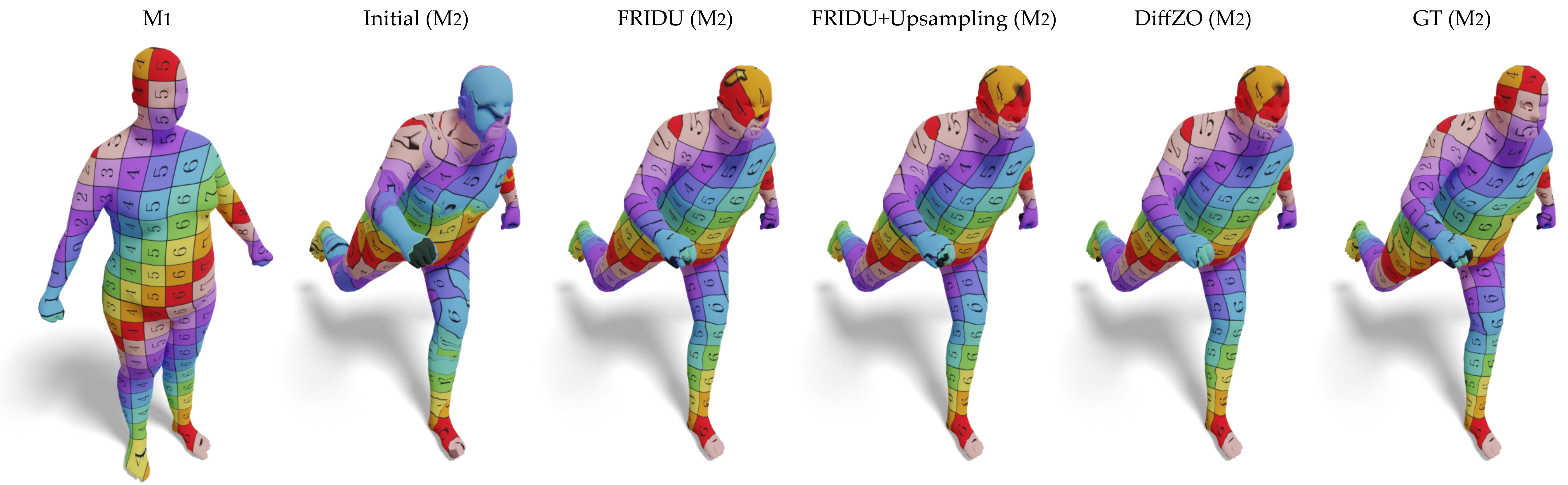}
    \caption{
    \change{
    A qualitative example comparing the map improvement obtained by our approach and by DiffZo~\cite{magnet2024memory}. For a pair of shapes from SHREC19, we show the resulting maps when both methods were trained on Faust (third column in Table~\ref{table:comparisons}) We show, from left to right: $M_1$, the initial map, our result with orthogonality guidance (FRIDU), our result with with orthogonality guidance and upsampling, the DiffZo result and the ground truth. We note that the initial map is very noisy, and FRIDU with upsampling achieves a very similar result to DiffZo despite the difficult initialization.   
    }}
    \label{fig:texture_cmp_diffzo}
\end{figure*}

In Figure~\ref{fig:zoomout-comparison}, we compare our method to ZoomOut on the three network models presented in Sect.~\ref{subsec:analysis} (i.e. initial map computation using WKS, WKS+landmarks and SHOT), trained on the \textit{michael} shapes.
Leveraging the flexibility of our guidance framework, we experiment with incorporating spectral upsampling during inference -- that is, progressively increasing the dimension of the functional map used to compute the pointwise correspondence throughout the denoising process. With upsampling our method performs comparably to ZoomOut in terms of accuracy. Furthermore, since the \textit{michael} meshes consist of approximately 50K vertices, the computational cost of ZoomOut is quite high ($\sim 600$ seconds per pair), while our inference procedure takes only $\sim 60$ seconds.

\change{
Figure~\ref{fig:texture_cmp_diffzo} shows a qualitative comparison between our approach and DiffZo when training on Faust and generalizing to SHREC19. We show texture transfer by transferring the texture coordinates generated on $M_1$ to the other meshes using the corresponding point-to-point map. Since all the maps are vertex to vertex, which leads to highly noisy texture transfer, we first apply a single iteration of RHM~\cite{ezuz2017deblurring} to all the maps. We note that the initial map that both methods start from is quite noisy, and our approach (with up-sampling) and DiffZo achieve comparable results. It is possible that a better initialization method may lead to improved results.
}

\subsection{Ablation}
\label{subsec:ablation}
\paragraph*{Zero-Shot Condition Generalization}
\begin{figure*}[t]
    \centering
    \includegraphics[width=\linewidth]{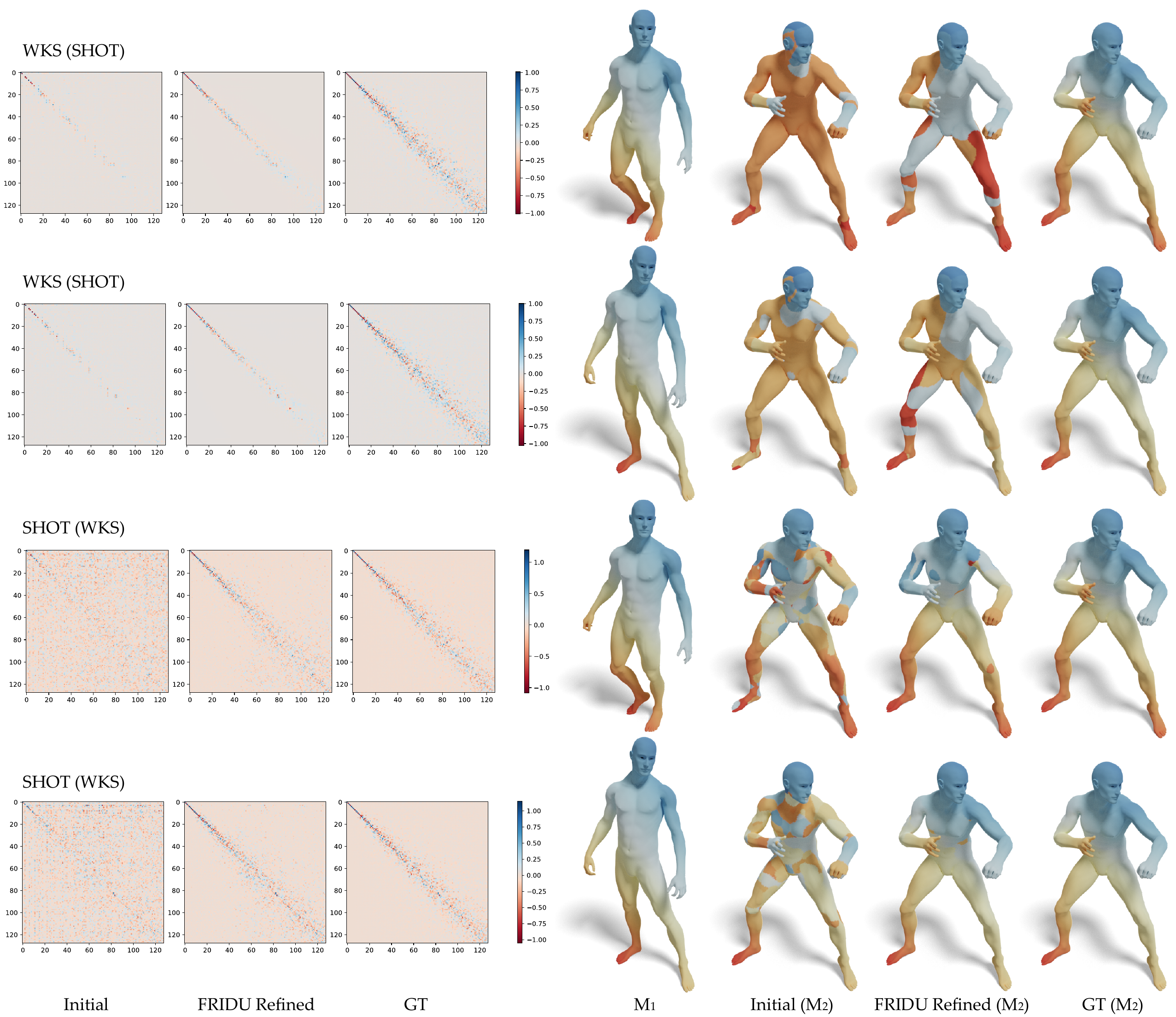}
    \caption{\textbf{Zero-Shot Condition Generalization} We test zero-shot condition generalization of our model by refining SHOT-based maps using a model trained on WKS-based maps, and vice versa, without additional training.
    The top two rows show the refinement of WKS-based maps using the SHOT-trained model, denoted as WKS (SHOT), while the bottom two rows show the refinement of SHOT-based maps using the WKS-trained model, denoted as SHOT (WKS).
    We observe that our model refines in both cases, with SHOT (WKS) achieving better performance than WKS (SHOT).
}
    \label{fig:michael-wks-shot-ablation}
\end{figure*}
We test our model's generalization ability to different sources of initial maps by refining WKS-based initial maps using the model trained on SHOT-based mappings, and vice versa.
Figure~\ref{fig:michael-wks-shot-ablation} shows two examples of WKS-based maps refined using the SHOT-based trained model (Section~\ref{subsec:shot-based-fm}), and two examples of SHOT-based maps refined using the WKS-based trained model (Section~\ref{subsec:analysis}). 
For each example, we show the pointwise mapping of \( f_1 \) and the corresponding functional map matrices.
We note that in both cases, our model manages to refine the initial map, but the results of refining SHOT-based maps using the WKS-based model are significantly better.

\paragraph*{Guidance Parameters}
\begin{figure*}[t]
    \centering
    \includegraphics[width=0.8\linewidth]{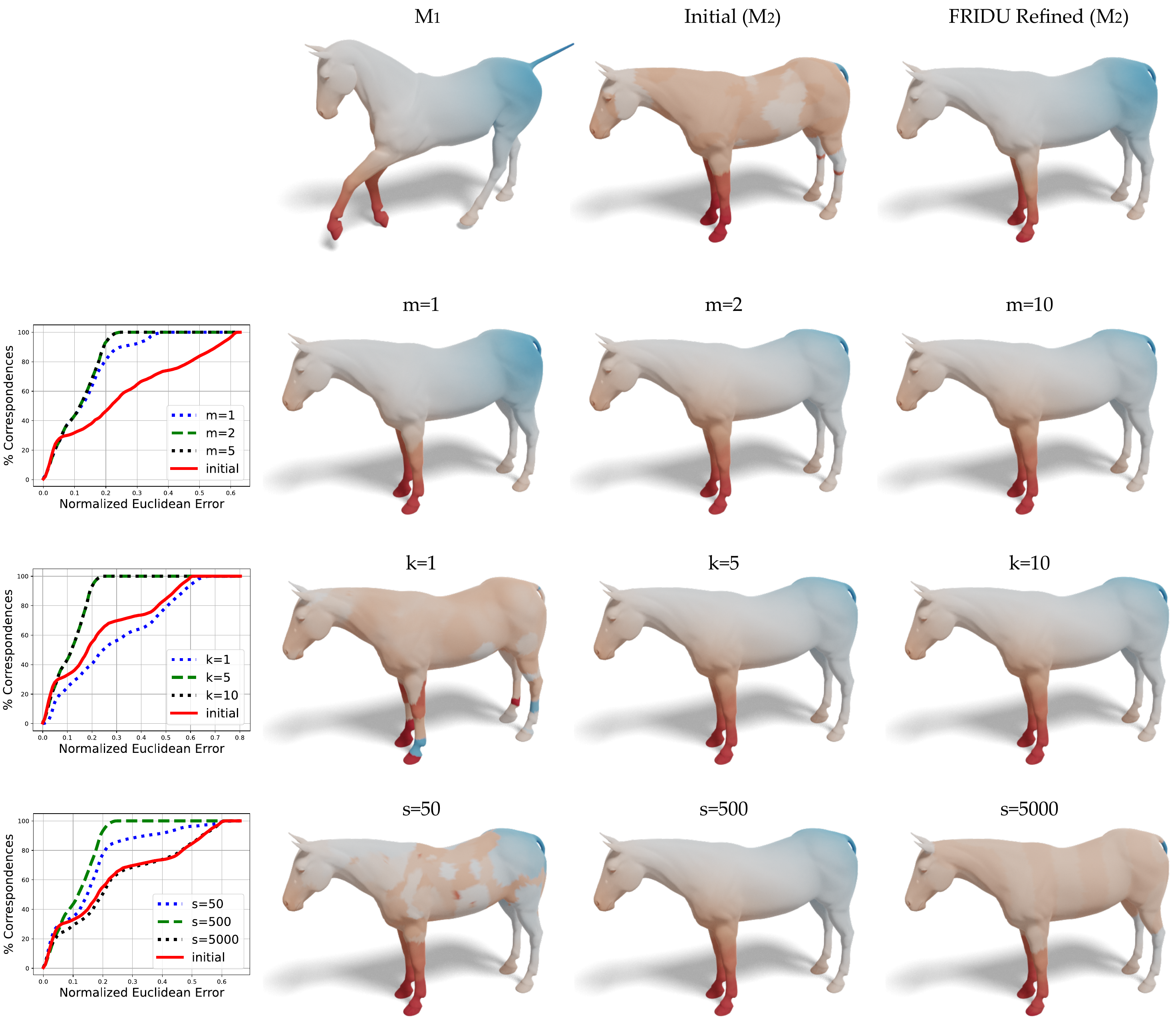}
    \caption{\textbf{Guidance Parameters.}
    Ablation study of the guidance parameters $m$, $k$, and $s$.  
    Top row: the source function on $\mathcal{M}_1$ and its mapping using the initial map and our refined map with our chosen parameters: $m = 2$, $k = 5$, $s = 500$.  
    In the three subsequent rows, we vary each parameter independently while keeping the other two fixed. The plots on the left show the corresponding normalized Euclidean error.}
    \label{fig:ablation-horse}
\end{figure*}
\begin{table}[b]
\centering
\begin{tabular}{|c|c|c|c|c|}
\hline
$m$ & $k$ & $s$ & $\downarrow$ Inference (s) & $\downarrow$ Mean Error \\
\hline
\textcolor{blue}{2} & \textcolor{blue}{5} & \textcolor{blue}{500} & \textcolor{blue}{16.20} & \textcolor{blue}{\textbf{0.10}} \\
2 & 10 & 500 & 31.72 & \textbf{0.10} \\
10 & 5 & 500 & 45.70 & \textbf{0.10} \\
1 & 5 & 500 & 9.18 & 0.12 \\
2 & 5 & 50 & 16.29 & 0.15  \\
2 & 5 & 5000 & 16.19 & 0.23 \\
2 & 1 & 500 & \textbf{3.68} & 0.28 \\
\hline
\end{tabular}
\caption{\textbf{Guidance Parameters.} Inference time and mean error for different values of \change{the guidance parameters $m$, $k$, and $s$ from ~\cite{bansal2023universal}}. The best result in each column is shown in bold. We note that our chosen parameters (first row) provide a good balance between inference time and accuracy.}
\label{table:ablation-horse}
\end{table}
We perform an ablation study on \change{the parameters of the guidance algorithm proposed in~\cite{bansal2023universal}. These are: the number of gradient steps for backward guidance \( m \), the number of recurrent steps \( k \), and the guidance strength parameter \( s \)}. We refer to the paper for more details. 
In all our experiments, we set \( m = 2 \), \( k = 5 \), and \( s = 500 \). For cross-dataset evaluations in Section~\ref{subsec:comparisons}, we find that increasing \( s \) to 2000 improves performance.

Figure~\ref{fig:ablation-horse} shows the performance of the model when inference is performed with different values of these guidance parameters. 
We use the model from Section~\ref{subsec:analysis}, trained on WKS-based functional maps between \textit{michael} shape pairs, and evaluate it on WKS-based functional map between \textit{horse} shape pairs from TACO.
The top row shows \( f_1 \), and the initial and our refined pointwise mapping using the default parameters \( m = 2, k = 5, s = 500 \). 
The next three rows show ablations for \( m \), \( k \), and \( s \), each with the other two parameters fixed.
Table~\ref{table:ablation-horse} provides the inference time in seconds and the normalized Euclidean error for each set of parameters. 
In general, we observe that increasing \( m \) and \( k \) increases inference time but improves performance only up to a point. The parameter \( s \) exhibits a balance point, where values that are too low or too high degrade performance. 
Overall, we find that our chosen parameters achieve a good trade-off between inference time and error in this example.

\change{
\section{Limitations}
As a refinement method, our approach is somewhat sensitive to the initial map. For example, as we show in Figure~\ref{fig:texture_cmp_diffzo}, given a very bad initialization our approach, while improving the map, does not fully reach the optimal solution. This is further exacerbated by training on FAUST and testing on SHREC19, which leads to additional generalization challenges, as can be seen in Table~\ref{table:comparisons}. In addition, we achieve only comparable performance to ZoomOut, when starting from the same initial map. However, in a setup where \emph{many} maps between meshes from the same class need to be improved (e.g., if we want to compute all the pairwise maps for a large dataset of meshes), our approach can offer a considerable speedup by first training on a subset of the dataset, and then efficiently testing on the rest of the shapes. Finally, we have only experimented with \emph{square} functional maps, whereas in practice one may prefer to use rectangular maps (e.g., using more basis functions on the target than on the source). We note that our algorithm extends naturally to this case, and all the guidance objectives can be used as-is. Yet we leave further validation to future work.
}

\section{Conclusion and Future Work}

We introduced \textbf{FRIDU}, a novel framework for refining functional maps via guided diffusion in the spectral domain. By interpreting functional maps as images, FRIDU leverages the generative capacity of diffusion models while enabling inference-time guidance to steer generation toward task-specific geometric properties. These objectives are applied modularly at test time, without requiring supervision or retraining, leading to high-quality correspondence refinement across datasets and diverse sources of initial maps.

We believe that diffusion models can be useful in additional aspects of the functional framework. For example, shape difference operators~\cite{rustamov2013map} and functional vector fields~\cite{azencot2013operator} are also operators which are potentially amenable to similar treatment using image diffusion models. Further, a composition of multiple maps (or of self-maps, generated by a tangent vector field) could potentially be analyzed using tools for videos.

Looking ahead, a major direction is to move beyond training a separate model for each type of input map. Instead, we envision \emph{foundation models~\cite{Bommasani2021FoundationModels} for functional maps}: diffusion-based priors trained on broad distributions of noisy correspondences that generalize across shape categories, datasets, and correspondence pipelines. Such models would support \emph{zero-shot refinement} via guidance alone, inspired by plug-and-play techniques developed in the diffusion literature, enabling modular adaptation without retraining. \textbf{FRIDU lays the groundwork for this direction} by demonstrating that flexible, loss-based guidance in the spectral domain is highly effective. To realize this vision, new forms of guidance may be required. One promising direction is to move beyond hand-crafted losses and explore \emph{learned guidance} techniques such as neural critics or reward models~\cite{fan2023dpok}, which could facilitate high-level constraints.

By connecting spectral geometry with modern generative modeling, FRIDU establishes a foundation for such cross-domain exploration. We hope this bridge between two previously unrelated paradigms will spur future work at the intersection of geometric analysis and diffusion-based generative modeling.

\paragraph*{Acknowledgments.}
Avigail Cohen Rimon and Mirela Ben Chen acknowledge the support of the Israel Science Foundation (grant No. 1073/21). Or Litany is a Taub fellow and is supported by the Azrieli Foundation Early Career Faculty Fellowship.

\bibliographystyle{eg-alpha-doi} 
\bibliography{FRIDU.bib}       

\appendix
\section{Implementation Details and Network Parameters}
\label{sec:impl_details}
To enable training a diffusion model on a very limited dataset, we adopt the Patch-Diffusion paradigm from \cite{wang2023patch} and build upon the code they provide.
Specifically, we use their implementation based on the UNet-based diffusion model EDM-DDPM++ \cite{karras2022elucidating}.
Following \cite{wang2023patch}, we employ the EDM sampling strategy with 50 deterministic reverse steps during inference.

In addition to default settings, we use the following hyperparameters for the Patch-Diffusion model:
$\text{num\_blocks} = 2$,
$\text{model\_channels} = 16$,
$\text{channel\_mult} = [2, 4]$,
and $\text{channel\_mult\_emb} = 0.1$.
We train using 3 patch resolutions, where the functional maps have dimension $128 \times 128$ in all experiments—except in Section 4.2, where we match the resolution used in DiffZO which is $130\times130$.
The parameter $\text{duration} = 2$ is used across all experiments, except in Section 4.2, where we set $\text{duration} = 10$ for the model trained on FAUST, and $\text{duration} = 15$ for the model trained on FAUST+SCAPE.

We use a batch size of 32 for all experiments, except for training on FAUST+SCAPE, where we use a batch size of 64.
We run all experiments on a single NVIDIA A40 GPU, except for the FAUST+SCAPE training, which we run on NVIDIA L40S GPU.

For guidance, we implement the algorithm described in \cite{bansal2023universal} with adaptation to the EDM.

\section{Initial Functional Map Computation}
\label{sec:init_fmap}
We use the \texttt{pyFM} Python package \cite{magnet2021pyfm} to generate initial functional maps for the experiments in Section 4.1, which also includes the computation of WKS descriptors. For the WKS we use the parameters $n\_descr = 100$, $subsample\_step = 5$ based on 150 eigenvectors.
For SHOT descriptors, we compute them in advance using the code from \cite{halimi2019unsupervised}, with the parameters $\text{num\_evecs} = 150$, $\text{num\_bins} = 10$, and $\text{radius} = 15$.

\section{DiffZO Comparison}
\label{sec:diffzo_cmp}
For comparison with the DiffZo method \cite{magnet2024memory}, we extract the weights of their pretrained Feature Extractor model (DiffusionNet \cite{sharp2022diffusionnet}) for models trained on FAUST and FAUST+SCAPE.
To compute the initial functional map used in our model, we first compute the pointwise map $\Pi_{\text{init}}$ from the extracted features and then derive the corresponding functional map using the corresponding Laplace-Beltrami eigenbases.

\section{Visualizations}
\label{sec:vis}
For visualizing functional map matrices, we use the colormap defined in \cite{liu2019spectral}, and for rendering in Blender, we utilize BlenderToolbox \cite{Liu_BlenderToolbox_2018}.

\end{document}